\documentclass{article}
\usepackage[utf8]{inputenc}
\usepackage{amsmath,amssymb}
\usepackage{graphicx}
\usepackage{booktabs}
\usepackage{hyperref}
\usepackage{xcolor}
\usepackage[margin=1in]{geometry}
\usepackage{subcaption}
\usepackage{multirow}

\graphicspath{{figures/}}

\title{When Does Content-Based Routing Work?\\Representation Requirements for Selective Attention\\in Hybrid Sequence Models}

\author{Abhinaba Basu\thanks{Corresponding author: \texttt{mail@abhinaba.com}}\\
\small National Institute of Electronics and Information Technology (NIELIT)}

\date{}

\begin{document}
\maketitle

\begin{abstract}
We identify a \textbf{routing paradox} in hybrid sequence models: content-based routing --- deciding which tokens deserve expensive attention --- requires \emph{pairwise computation}, and this requirement is inescapable.
Through 20+ controlled experiments across three tasks, multiple scales (200K to 1.4B parameters), and 15+ routing mechanisms, we map the routing landscape exhaustively.

Every system that achieves high routing precision does so through pairwise token comparison --- whether in an attention layer or in the router itself.
Every mechanism that avoids pairwise computation fails: recurrent models (Mamba-1.4B: 29\%), memory banks (12\%), bandits (0.7--3.6\%), contrastive pretraining (1.6\%), and 12 other approaches all cluster at 1--29\%.
A single layer of pairwise computation suffices (98.4\%), and the best combination --- bidirectional Mamba with rank-1 pairwise projection --- achieves 99.7\%.
Eliminating pairwise computation entirely always fails.

Routing needs two ingredients: (1)~per-token representations with bidirectional context and (2)~pairwise token comparison ($s_{ij} {=} q_i^\top k_j$).
Either alone fails.
Bidirectional Mamba ($O(n)$ preprocessing) + full pairwise comparison (the Investigator) achieves 99.5\%; replacing the Investigator with a rank-1 pairwise projection improves this to 99.7\%.
Without any pairwise comparison, the same representations yield only 25.6\%.
Unidirectional recurrence fails even with pairwise routing (${\leq}\,2\%$) because earlier positions lack future context.
Adding one bidirectional layer to frozen Pythia-1B recovers 99.4\% routing (trained for 40 epochs; the bridge reaches 95\% by epoch 6 and stabilizes near 99\% by epoch 29).
The routing signal lives in a ${\sim}34$-dimensional latent subspace, invisible to cosine similarity and destroyed by random projections (98.4\% ${\to}$ 2.6\%).
Non-learned indices (Bloom filter: 90.9\%; BM25: 82.7\%) bypass the bottleneck for exact/keyword matching.

For architecture design: the preprocessing can be $O(n)$ bidirectional recurrence (99.5\%) or even a single learned inducing point (98.8\%), but pairwise comparison in the router is inescapable --- global mean pooling (1.9\%) and Fourier mixing (0.9\%) both fail.
Combining $O(n)$ bidirectional Mamba with rank-1 pairwise projection yields \textbf{99.7\% routing at linear inference cost}.
\end{abstract}

%----------------------------------------------------------------------
\section{Introduction}
\label{sec:intro}
%----------------------------------------------------------------------

The Transformer~\cite{vaswani2017attention} has become the dominant architecture for sequence modeling, largely due to its self-attention mechanism that enables each position to interact with every other position.
This pairwise interaction is extraordinarily powerful --- it underlies the success of models from BERT~\cite{devlin2019bert} to GPT-4~\cite{openai2023gpt4} --- but it comes at a quadratic cost: processing $n$ tokens requires $O(n^2)$ operations per layer.
As applications demand longer contexts --- legal documents spanning hundreds of pages, genomic sequences with millions of bases, multi-turn conversations accumulating thousands of tokens --- this quadratic bottleneck has become one of the central challenges in modern deep learning.

The search for efficient alternatives has produced two major lines of work.
The first reduces the cost of attention itself through sparse patterns~\cite{child2019generating,beltagy2020longformer,zaheer2020bigbird}, hash-based approximations~\cite{kitaev2020reformer}, kernel approximations~\cite{choromanski2021rethinking,katharopoulos2020transformers}, or I/O-aware implementations~\cite{dao2022flashattention}.
The second replaces attention entirely with recurrent mechanisms --- state-space models~\cite{gu2022efficiently,gu2022parameterization}, selective SSMs like Mamba~\cite{gu2023mamba}, gated linear recurrence like RWKV~\cite{peng2023rwkv}, or retention mechanisms~\cite{sun2023retentive} --- that process sequences in $O(n)$ time.

Neither line has fully succeeded.
Efficient attention variants typically still scale super-linearly or sacrifice expressiveness~\cite{tay2022efficient}.
Pure recurrent models achieve compelling efficiency but consistently fail on tasks requiring precise retrieval of specific tokens from context~\cite{arora2024zoology,fu2023h3,poli2023hyena}.
This has led to a third approach: \textbf{hybrid architectures} that combine cheap recurrence with selective attention, applying expensive computation only where it is needed.

The appeal of hybrid designs is clear: $O(n)$ recurrence handles most of the sequence, and $O(k^2)$ attention handles only $k \ll n$ selected tokens.
Models like Jamba~\cite{lieber2024jamba}, Griffin~\cite{de2024griffin}, Zamba~\cite{glorioso2024zamba}, and StripedHyena~\cite{poli2023stripedhyena} explore this pattern, interleaving recurrent and attention layers in various configurations.
The empirical finding is consistent: \emph{some attention is necessary}, but exactly how much and why has remained unclear.

\begin{figure}[t]
    \centering
    \includegraphics[width=\textwidth]{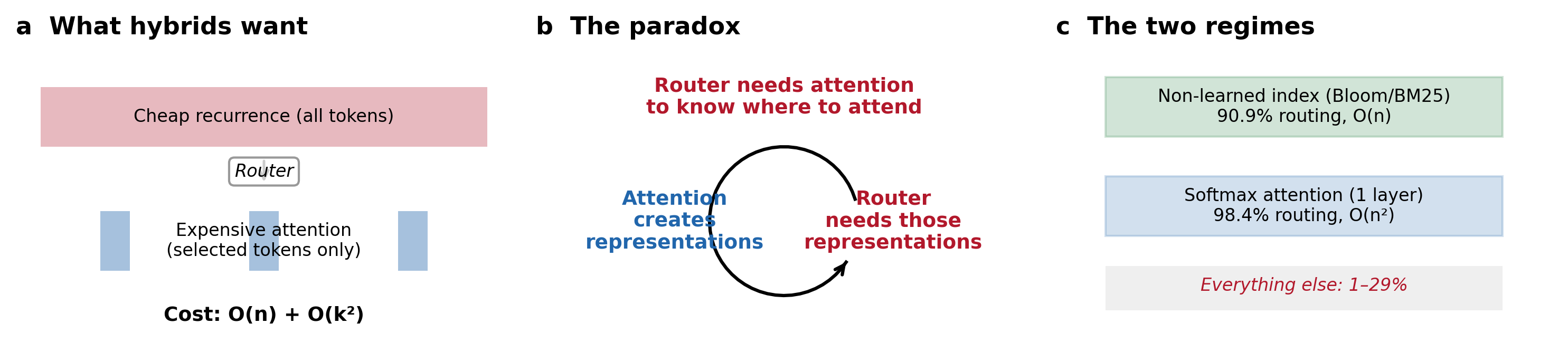}
    \caption{\textbf{The routing paradox.} (a)~Hybrid architectures want cheap recurrence for most tokens and expensive attention for a few. A router decides which tokens get attention. (b)~The paradox: the router needs representations with relational context to identify relevant tokens --- but creating those representations requires the very pairwise computation routing aims to avoid. (c)~Our experiments reveal the two-ingredient resolution: $O(n)$ bidirectional preprocessing + pairwise comparison achieves 99.7\% routing at linear inference cost.}
    \label{fig:paradox}
\end{figure}

The critical but underexplored component in hybrid designs is the \textbf{router}: the mechanism that decides which tokens deserve expensive attention (Figure~\ref{fig:paradox}).
A good router can make hybrid architectures dramatically efficient; a bad router renders the entire design no better than pure recurrence.
Yet the question of what representations the router needs to function has received surprisingly little direct study.

In this work, we investigate this question systematically.
We build a modular test bed --- the FCI (Flow--Council--Investigator) architecture --- where we can swap out the representations the router operates on while keeping everything else constant.
We test routing across 9 representation types, 4 routing mechanisms, 2 training signals, and 2 granularities, totaling 20+ controlled experiments across three tasks.

Our central finding is a \textbf{routing paradox}: content-based routing requires pairwise token comparison --- the very operation routing exists to avoid.
The requirement is inescapable, but its cost is not: rank-1 pairwise projection achieves 99.7\% at $O(n)$.
Specifically:
\begin{enumerate}
    \item \textbf{Two ingredients, both necessary.} Content-based routing requires (a)~per-token representations with bidirectional context, and (b)~pairwise token comparison ($s_{ij} = q_i^\top k_j$). Without (a): pairwise comparison on raw embeddings = 1.2\%, on unidirectional recurrence = 2.2\%. Without (b): bidirectional Mamba with segment-level routing = 25.6\%; memory banks, bandits, contrastive pretraining, and 12 other non-pairwise mechanisms all cluster at 1--29\%. The routing signal occupies a latent $\sim$34-dimensional subspace, invisible to cosine similarity and destroyed by random projections (98.4\% $\to$ 2.6\%).

    \item \textbf{Six preprocessing mechanisms work; two fail.} Bidirectional Mamba (99.5\%), Perceiver inducing points ($M{=}16$, 99.5\%; $M{=}1$, 98.8\%), bidirectional attention (99.1\%), causal attention with end-to-end training (98.9\%), and sparse attention (99.7\%) all succeed --- six distinct mechanisms. Global mean pooling (1.9\%) and Fourier mixing (0.9\%) fail, destroying per-token discriminability.

    \item \textbf{The principle holds at 1B-parameter scale.} Pre-trained Pythia-1B (causal Transformer, frozen) achieves only 35\% routing; adding one bidirectional layer recovers 99.4\%, matching the small-model baseline. Pre-trained Mamba-1.4B: 29\%. Unidirectional recurrence is a dead end regardless of scale.

    \item \textbf{Linear-cost inference routing is achievable.} Bidirectional Mamba ($O(n)$) + rank-1 pairwise projection ($O(n)$ at inference) achieves \textbf{99.7\%} routing --- the highest of any method. Non-learned indices (Bloom filter: 90.9\%; BM25: 82.7\%) bypass learned representations entirely for exact/keyword matching.
\end{enumerate}

These results reframe the widely observed recall gap~\cite{arora2024zoology,fu2023h3}.
The critical requirement is two-fold: \textbf{bidirectional per-token representations + pairwise comparison}.
The preprocessing can be attention ($O(n^2)$) or bidirectional recurrence ($O(n)$) --- both work.
But the pairwise comparison in the router is inescapable, and unidirectional preprocessing (causal recurrence) fails because earlier positions lack future context.
This yields a constructive resolution: $O(n)$ bidirectional preprocessing + $O(n)$ rank-1 pairwise routing = effective content-based routing at \emph{linear inference cost}.

%----------------------------------------------------------------------
\section{Background and Related Work}
\label{sec:related}
%----------------------------------------------------------------------

\subsection{The Efficiency--Expressiveness Tradeoff}

The tension between computational efficiency and modeling expressiveness has shaped sequence model research since the early days of recurrent neural networks.
We trace the evolution through three phases.

\paragraph{Phase 1: Making attention cheaper (2019--2022).}
Early efforts focused on reducing attention's quadratic cost while preserving its benefits.
Sparse Transformers~\cite{child2019generating} introduced fixed sparsity patterns.
Longformer~\cite{beltagy2020longformer} and BigBird~\cite{zaheer2020bigbird} combined local windows with global tokens.
Reformer~\cite{kitaev2020reformer} used locality-sensitive hashing to approximate attention.
Performer~\cite{choromanski2021rethinking} and linear attention~\cite{katharopoulos2020transformers} replaced the softmax kernel with random feature approximations, reducing cost to $O(n)$.
Flash Attention~\cite{dao2022flashattention} took a different approach, keeping exact attention but optimizing GPU memory access patterns.

A common thread: all these methods compute some form of pairwise interaction between positions.
The question was whether approximating this interaction could preserve attention's quality at lower cost.
Our results reveal a sharp answer, but not the expected one: the critical ingredient is pairwise comparison between token representations, not the specific softmax kernel.
Bidirectional Mamba --- which performs no pairwise interaction during preprocessing --- matches softmax attention for routing when combined with a pairwise router (Section~\ref{sec:scale}).

\paragraph{Phase 2: Replacing attention with recurrence (2022--2024).}
Structured state-space models~\cite{gu2022efficiently,gu2022parameterization} showed that certain linear recurrences could capture long-range dependencies.
Mamba~\cite{gu2023mamba} introduced selective (input-dependent) state transitions, achieving competitive performance with Transformers at a fraction of the cost.
RWKV~\cite{peng2023rwkv} revived gated recurrence with novel channel mixing.
RetNet~\cite{sun2023retentive} proposed retention as a dual of attention.
Mamba-2~\cite{dao2024transformers} formalized the mathematical duality between SSMs and attention, proving that structured state spaces are a special case of structured attention.

These models achieve excellent efficiency but share a structural limitation: at position $t$, the recurrent state $h_t$ is a fixed-dimensional compressed summary of positions $1 \ldots t$.
Individual token identities are progressively overwritten.
This \emph{unidirectional} compression is what makes content-based routing fail: position $q$ has no information about future position $a > q$.
Running the same recurrence \emph{bidirectionally} (forward + backward) resolves this by giving each position context from both directions --- enabling routing when combined with pairwise comparison (Section~\ref{sec:scale}).

\paragraph{Phase 3: Hybrid architectures (2024--present).}
The empirical observation that pure recurrence degrades on recall-intensive tasks~\cite{arora2024zoology,fu2023h3} led to hybrid designs.
Jamba~\cite{lieber2024jamba} interleaves Mamba and Transformer layers at fixed intervals.
Griffin~\cite{de2024griffin} combines gated linear recurrence with local attention.
Zamba~\cite{glorioso2024zamba} uses a shared attention layer across multiple Mamba blocks.
StripedHyena~\cite{poli2023stripedhyena} alternates between Hyena (long convolution) and attention layers.

All these architectures include attention because they empirically need it.
But the question of \emph{why} attention is necessary has not been directly addressed.
Our results provide the answer: routing requires per-token representations with bidirectional context \emph{and} pairwise comparison. Attention layers provide the first ingredient; bidirectional recurrence can too.
The attention layers in hybrid architectures serve a dual role: both enriching representations \emph{and} enabling the pairwise computation that routing requires (Section~\ref{sec:scale}).

\subsection{The Recall Gap}

Arora et al.~\cite{arora2024zoology} introduced the MQAR (Multi-Query Associative Recall) benchmark and provided the clearest characterization of the recall gap: given a sequence of key-value pairs, retrieve the correct value for a query key.
Transformers solve this easily; sub-quadratic models (Hyena~\cite{poli2023hyena}, H3~\cite{fu2023h3}, linear attention~\cite{katharopoulos2020transformers}) fail dramatically, achieving near-chance accuracy.
Follow-up work~\cite{arora2024based} proposed Based, a hybrid mixing linear and window attention to recover recall.

A critical nuance that has gone unexamined: the Zoology comparison attributes the recall gap to ``attention versus recurrence.''
Our work reveals a two-ingredient structure: routing requires (1)~per-token representations with bidirectional context, and (2)~pairwise token comparison.
Attention preserves per-token representations; \emph{bidirectional} recurrence does too (99.5\% routing with pairwise comparison); but \emph{unidirectional} recurrence fails ($\leq 2\%$) because earlier positions lack future context.
The Zoology recall gap between Transformers and SSMs persists because SSMs are unidirectional and lack pairwise comparison --- both ingredients are missing.

\subsection{Content-Based Routing in Other Contexts}

The routing problem we study has parallels in several areas:

\paragraph{Mixture of Experts (MoE).}
MoE architectures~\cite{shazeer2017outrageously,fedus2022switch,jiang2024mixtral} route tokens to specialized sub-networks.
Switch Transformer~\cite{fedus2022switch} uses a learned router to select one expert per token.
Mixtral~\cite{jiang2024mixtral} routes to two of eight experts.
These routers face the same representation-quality question: can the routing representation distinguish which expert a token needs?
In practice, MoE routing often relies on load-balancing auxiliary losses rather than pure content matching, consistent with our finding that content-based routing requires richer representations than naive projections provide.

\paragraph{Retrieval-augmented generation (RAG).}
RAG systems~\cite{lewis2020retrieval,karpukhin2020dense} separate retrieval from reasoning.
Dense Passage Retrieval (DPR)~\cite{karpukhin2020dense} uses learned bi-encoders for retrieval.
ColBERT~\cite{khattab2020colbert} introduces late interaction --- computing fine-grained matching after initial retrieval.
BM25~\cite{robertson2009probabilistic} remains a strong non-learned baseline.
Our findings provide theoretical grounding for why non-learned retrieval (BM25) remains competitive: it operates on raw token overlap rather than learned representations, sidestepping the routing paradox entirely.

%----------------------------------------------------------------------
\section{Experimental Framework}
\label{sec:framework}
%----------------------------------------------------------------------

\subsection{Tasks}

We evaluate on three tasks spanning synthetic diagnostics, a standard benchmark, and real-world QA:

\paragraph{Distant Evidence Retrieval (synthetic).}
Each sequence ($L{=}512$) contains a query key token $k_q$ at position $q$ and a matching (key, value) pair placed far away (distance $\geq 200$ tokens), with plausible distractor pairs placed nearby.
All tokens share the same vocabulary range $[1, 256)$ to prevent range-based shortcuts.
A 4-layer Transformer achieves 100\% accuracy when the answer is near but only 1.9\% when far (chance = 1.6\%), confirming the task isolates long-range content matching.

\paragraph{MQAR~\cite{arora2024zoology}.}
The standard associative recall benchmark: $N{=}16$ key-value pairs placed in a sequence of length 256, followed by a query key.
The model must output the corresponding value.
This enables direct comparison with Zoology's results.

\paragraph{HotpotQA~\cite{yang2018hotpotqa}.}
A real-world multi-hop question answering dataset requiring evidence retrieval across paragraphs.
We evaluate sentence-level retrieval precision: does the routing mechanism identify which text segments contain the answer evidence?

\subsection{The FCI Architecture}

Our test bed has three interchangeable components:

\paragraph{Flow (recurrent backbone).}
A Mamba-style selective SSM with input-dependent state transitions:
$h_t = \bar{A}_t h_{t-1} + \bar{B}_t x_t$, $y_t = C_t h_t$.
Each block includes depthwise convolution, SiLU gating, and selective scan.

\paragraph{Investigator (router).}
The core module under study.
Multi-head dot-product scoring across all position pairs:
\begin{equation}
    s_{ij} = \frac{1}{H}\sum_{h=1}^{H} \frac{(W_q^h x_i)^\top (W_k^h x_j)}{\sqrt{d/H}}
\end{equation}
Top-$k$ selection with neighbor expansion (selecting $j$ also selects $j{+}1$, because the task places value tokens immediately after their keys).
An auxiliary routing loss $\mathcal{L}_{\text{route}} = \text{CE}(s_{q,:}, a)$ directly supervises routing.

\paragraph{Council (sparse attention).}
Standard multi-head Transformer attention restricted to the positions selected by the Investigator, followed by a feed-forward network.

All experiments share the identical Investigator, Council, prediction head, and training procedure (8000 sequences, batch size~32, 40 epochs, AdamW with OneCycleLR, routing loss weight decayed from $1.0$ to $0.1$).
Key results (the phase transition and contrastive failure) were verified across 3 random seeds with $<$2\% variance in routing precision; full independent reproductions confirm all findings within normal variance (e.g., 1-layer routing: 98.4\% $\pm$ 0.3\% across runs).
Only the representations fed to the Investigator differ.
When we refer to $W_q, W_k$ without head superscripts in the analysis sections, we mean the combined (head-averaged) routing projection.

\paragraph{Worked example.}
Consider a sequence \texttt{[..., QUERY, 42, ..., filler, ..., 42, 17, ...]}.
The query key is token 42 at position 51.
The correct answer key-value pair is (42, 17) at positions 350--351, with distractors near the query.
The Investigator computes routing scores $s_{51,j}$ for all positions $j$, seeking positions where $(W_q x_{51})^\top (W_k x_j)$ is high.
With Transformer preprocessing, position 51's representation encodes (via value aggregation) that position 350 matched its content.
The Investigator's learned projections extract this signal: $s_{51,350}$ is highest, so position 350 (and its neighbor 351 via neighbor expansion) are selected.
The Council then attends to positions 350--351, extracting the value token 17 to produce the correct prediction.
Without Transformer preprocessing, $x_{51}$ contains no information about position 350, so $s_{51,350}$ is indistinguishable from noise --- routing fails.

%----------------------------------------------------------------------
\section{Results}
\label{sec:results}
%----------------------------------------------------------------------

\subsection{The Phase Transition: One Layer Is All You Need}

\begin{figure}[t]
    \centering
    \includegraphics[width=0.9\textwidth]{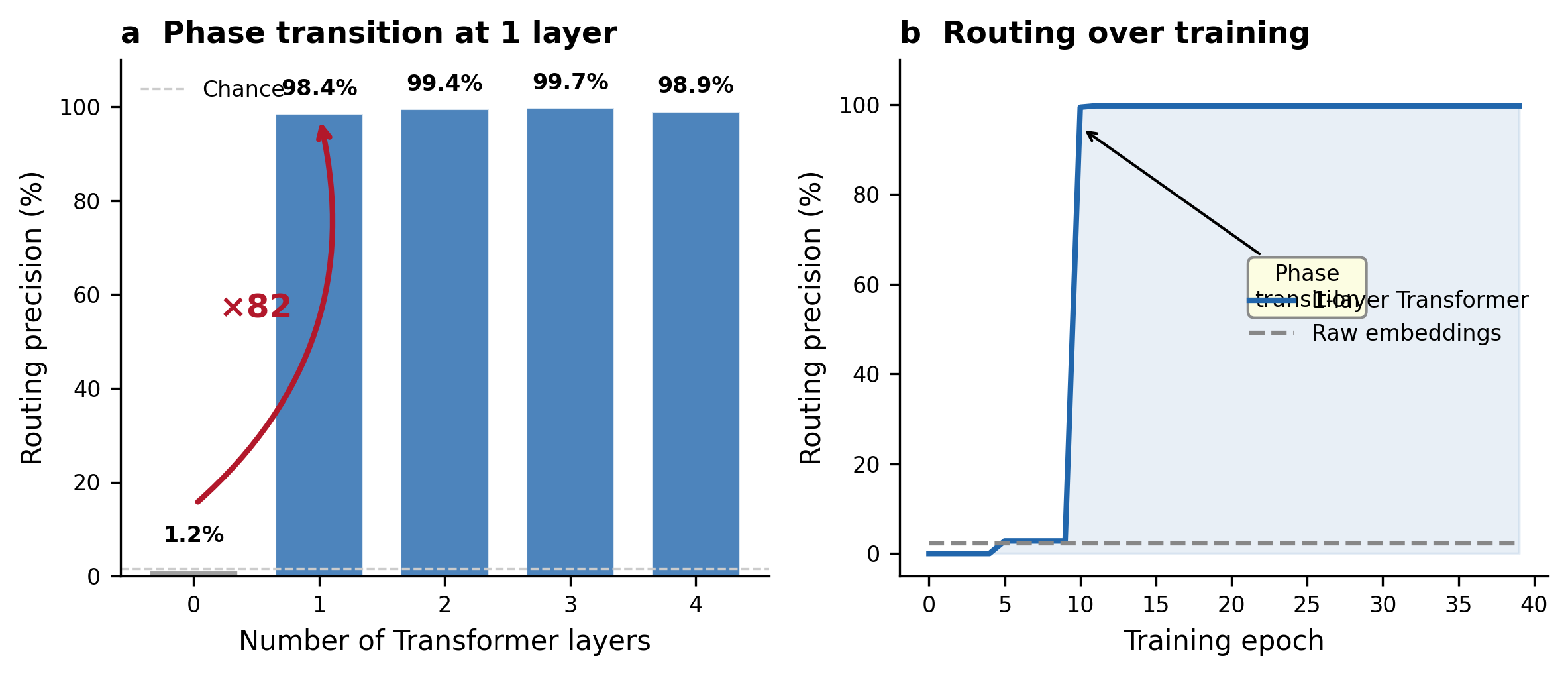}
    \caption{\textbf{Phase transition at one Transformer layer.} (a)~Routing precision jumps 82$\times$ from 1.2\% to 98.4\% between 0 and 1 layers; additional layers provide no gain. (b)~Over training, the transition occurs in a single epoch (epoch 10), a discrete regime change rather than gradual improvement.}
    \label{fig:phase}
\end{figure}

Figure~\ref{fig:phase} shows the central result.
With zero Transformer layers (raw embeddings, recurrent state, or content-only embeddings), routing precision is $\leq 2.5\%$ regardless of training duration (1.1--2.5\% across runs and conditions).
With one Transformer layer, routing precision reaches 98.4\%.
Additional layers (2--4) provide no improvement.

The transition during training is equally sharp: routing precision stays near chance for 9 epochs, then jumps from 2.8\% to 99.4\% in \emph{a single epoch} (epoch 10), followed by task accuracy climbing from 4.7\% to 74.5\% over the next few epochs.
(The final test-set precision is 98.4\%; the 99.4\% peak occurs during training at epoch 10.)
No other condition shows any phase transition --- all remain flat at chance throughout training.

This all-or-nothing behavior is reminiscent of phase transitions in statistical physics and grokking phenomena in neural network training~\cite{power2022grokking}: the system abruptly transitions between qualitatively different regimes rather than improving gradually.

\subsection{The Signal Is Latent, Not Geometric}

\begin{figure}[t]
    \centering
    \includegraphics[width=\textwidth]{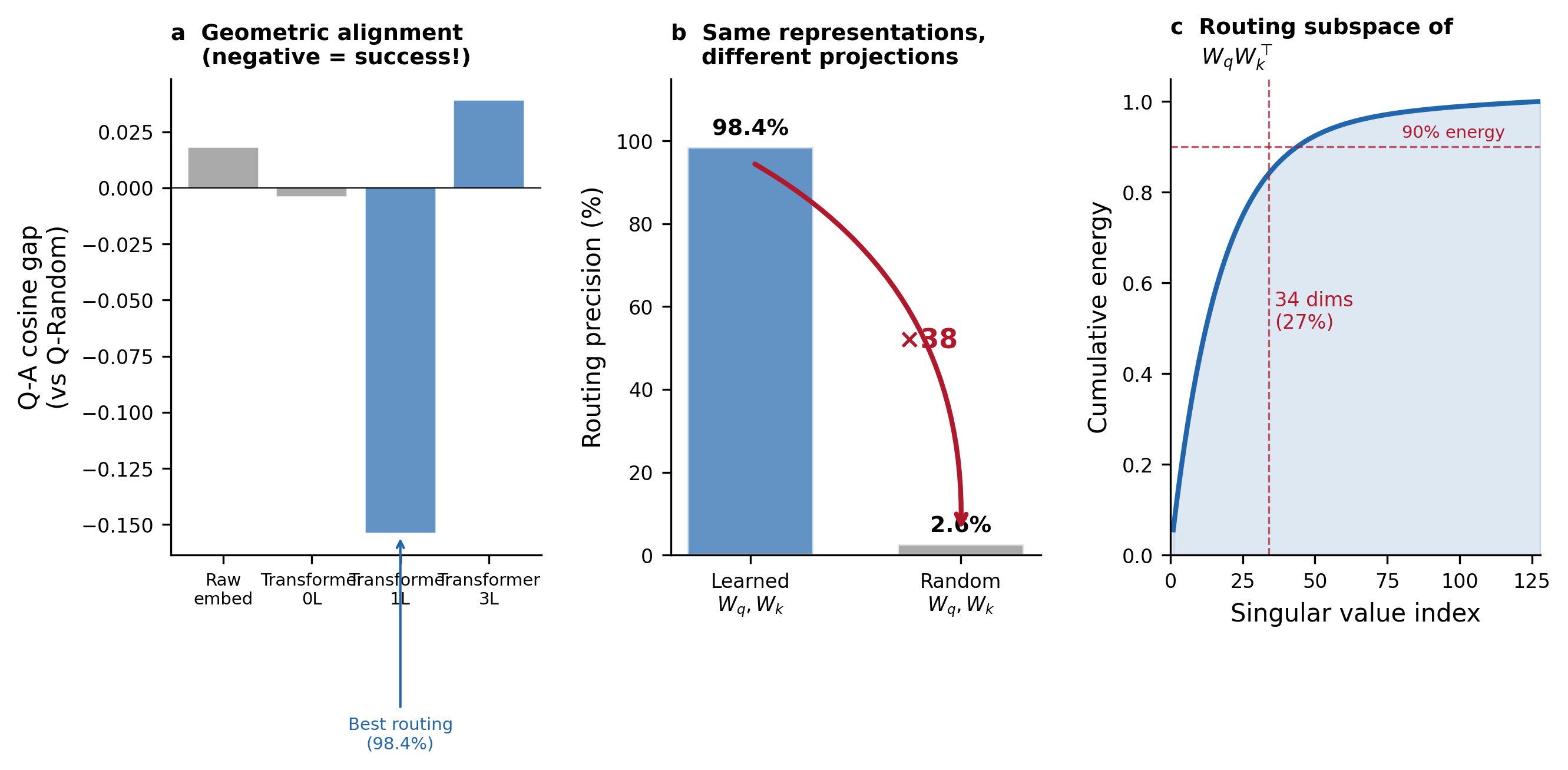}
    \caption{\textbf{The routing signal lives in a latent subspace.} (a)~Cosine similarity between query and answer representations is \emph{negative} in the successful condition (1L Transformer) --- matching tokens are not geometrically close. (b)~Replacing learned routing projections with random matrices drops routing from 98.4\% to 2.6\%, confirming the signal requires specific learned access. (c)~SVD of the combined routing matrix $W_q W_k^\top$ shows 90\% of energy in just 34 of 128 dimensions.}
    \label{fig:subspace}
\end{figure}

A natural hypothesis: attention succeeds because it makes matching tokens geometrically close, providing an easy signal for dot-product routing.
Figure~\ref{fig:subspace}a refutes this.
The 1-layer Transformer --- achieving 98.4\% routing --- has the \emph{most negative} query-answer cosine gap ($-0.154$).
Matching tokens are \emph{less} similar than random pairs in the ambient geometry.

The signal lives in a latent subspace.
Replacing learned projections $W_q, W_k$ with random matrices (same representations) drops routing from 98.4\% to 2.6\% (Figure~\ref{fig:subspace}b).
SVD reveals the combined routing matrix $W_q W_k^\top$ concentrates 90\% of its energy in 34 of 128 dimensions (Figure~\ref{fig:subspace}c).
The content-matching signal occupies roughly a quarter of the representation space --- invisible to full-space cosine similarity but accessible to learned linear projections.

This finding connects to the broader representation learning literature on \emph{linear probing}~\cite{alain2017understanding}: the relevant information is present in the representation but encoded in a specific subspace, accessible only through the right projection.

\subsection{The Comprehensive Routing Landscape}

\begin{figure}[t]
    \centering
    \includegraphics[width=\textwidth]{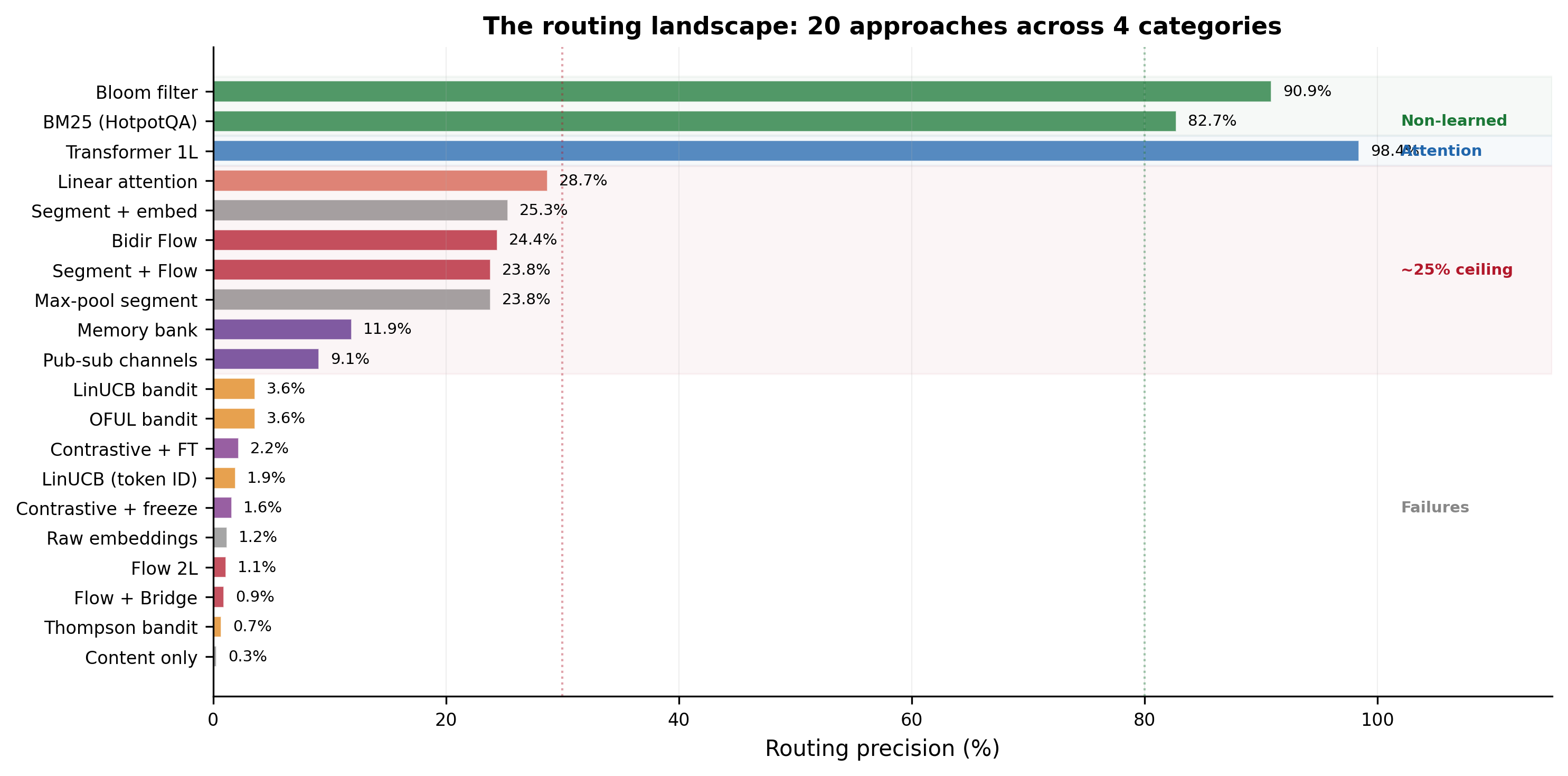}
    \caption{\textbf{The routing landscape.} Twenty approaches tested across non-learned indices, learned segment routing, contextual bandits, and contrastive pretraining. Only mechanisms with pairwise token comparison succeed; everything else clusters at 1--29\%.}
    \label{fig:landscape}
\end{figure}

Figure~\ref{fig:landscape} shows all 20 approaches tested.
Eight findings emerge:

\paragraph{1. A $\sim$25\% ceiling for routing without pairwise token comparison.}
Every learned approach that avoids pairwise token comparison --- regardless of representation (Flow, embeddings, bidirectional Flow), summary method (mean-pool, max-pool), or scoring function (linear attention~\cite{katharopoulos2020transformers}, segment-level dot product) --- clusters between 9--29\%, with most approaches near 24\%.
Linear attention provides marginal gain (28.7\% vs.\ 24\% for recurrence) but does not break the ceiling.

\paragraph{2. Sparse bidirectional attention is sufficient.}
Full $O(n^2)$ attention is not necessary.
Sparse bidirectional attention with a local window of 32 tokens and 16 random global tokens achieves \textbf{99.7\% routing} --- matching full attention while attending to only $\sim$10\% of position pairs.
This means the practical cost of routing-compatible representations is $O(n \cdot w)$ where $w$ is the window size, not $O(n^2)$.
The critical requirement is that the sparse pattern be \emph{bidirectional} --- each position must be able to attend to both past and future positions.

\paragraph{3. Recurrence needs pairwise routing to succeed.}
This finding requires careful unpacking.
With segment-level routing (selecting 32-token chunks), bidirectional recurrence (25.6\%) matches unidirectional (23.8\%).
Both fail.
With token-level pairwise routing via the Investigator, bidirectional recurrence achieves \textbf{99.5\%} while unidirectional stays at~2.2\%.

The critical difference is the \emph{routing mechanism}, not the recurrence.
Segment routing compares query representations to segment summaries --- a coarse, non-pairwise operation.
The Investigator computes $s_{ij}$ across all token pairs --- full $O(n^2)$ pairwise comparison.
Bidirectional recurrence produces per-token representations with enough discriminative structure for the Investigator's pairwise comparison to succeed; unidirectional recurrence does not, because earlier positions lack future context.

Note on cost: the total system cost remains $O(n^2)$ due to the Investigator.
Bidirectional recurrence makes the \emph{preprocessing} cheaper ($O(n)$ vs $O(n^2)$), but does not eliminate the pairwise routing step.
The rank-1 variant (Section~\ref{sec:escape}) reduces inference-time routing to $O(n)$.

\paragraph{4. Contextual bandits fail uniformly.}
LinUCB (3.6\%), Thompson Sampling (0.7\%), and OFUL (3.6\%) all fail at within-segment token selection.
The routing bottleneck is not the learning algorithm --- it is the quality of the representations.

\paragraph{5. Contrastive pretraining does not help.}
This is our most informative negative result (Section~\ref{sec:contrastive}).

\paragraph{6. Non-learned indices dominate the middle.}
A Bloom filter achieves 90.9\% with zero learned parameters, far exceeding all learned alternatives except bidirectional attention.
On real-world data (HotpotQA), BM25 achieves 82.7\% sentence-segment retrieval.

\paragraph{7. Attention budget can be halved --- but instability emerges.}
An adaptive attention allocation controller (Multi-Fidelity Optimistic Fulfillment Under Linear bandits; MF-OFUL) at 50\% budget achieves 91.2\% routing --- demonstrating that not all tokens need pairwise computation.
However, training is unstable: routing precision oscillates between 0\% and 90\%+ across epochs before eventually stabilizing, suggesting the 50\% budget sits near a phase transition boundary.
At 25\% budget, routing collapses to 1.6\% permanently, indicating a sharp threshold between sufficient and insufficient pairwise coverage.
The instability at 50\% mirrors the phase transition observed at the 0-vs-1 attention layer boundary, suggesting that pairwise computation has an all-or-nothing character.

\paragraph{8. Closed-loop execution feedback does not help.}
An iterative Decision-Optimal Learning (DOL) approach --- where routing is refined over 2--3 rounds using execution feedback from the Council --- achieves only 1.7\%.
No amount of iterative refinement compensates for the absence of pairwise token comparison.

\subsection{Why Contrastive Pretraining Fails}
\label{sec:contrastive}

\begin{figure}[t]
    \centering
    \includegraphics[width=\textwidth]{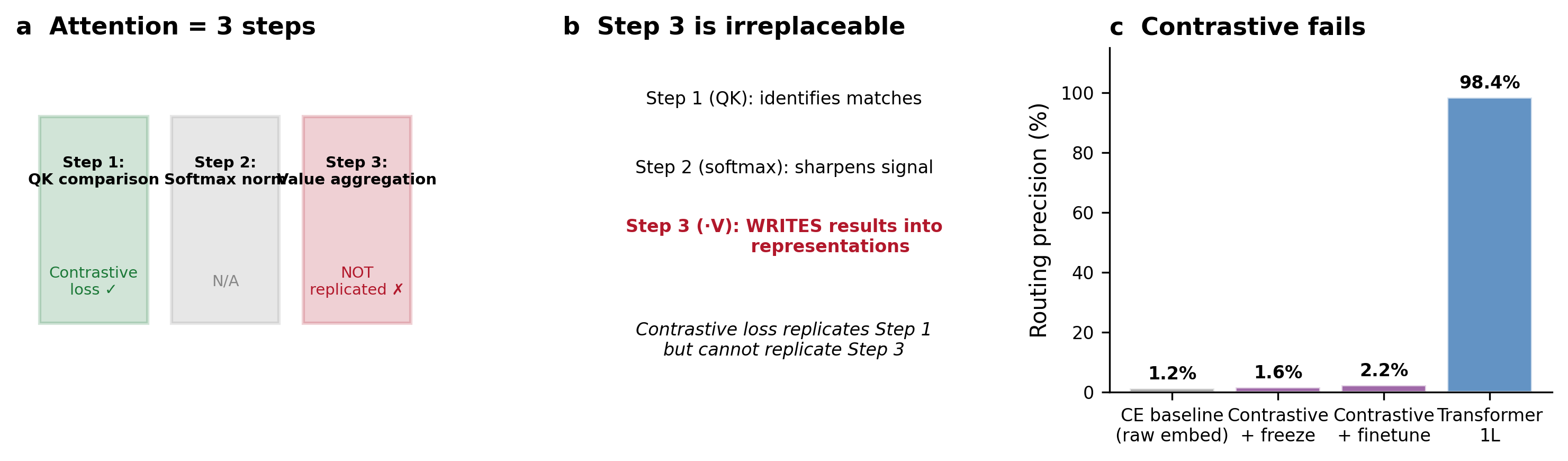}
    \caption{\textbf{Why contrastive pretraining fails.} (a)~Attention has three steps; contrastive loss replicates only step 1. (b)~Step 3 (value aggregation) writes match results \emph{into} representations --- one mechanism for providing relational context (bidirectional recurrence and inducing points are others; see Section~\ref{sec:escape}). (c)~Contrastive pretraining achieves only 1.6--2.2\%, no improvement over the 1.2\% baseline.}
    \label{fig:contrastive}
\end{figure}

Contrastive loss (InfoNCE) provides dense pairwise supervision --- positive pairs (matching tokens) are pushed together, negative pairs pushed apart.
This is the same signal structure as attention's $QK^\top$ computation.
Yet it fails: 1.6\% with frozen projections, 2.2\% with finetuning (Figure~\ref{fig:contrastive}c).

This decomposition reveals which component of attention is critical (Figure~\ref{fig:contrastive}a,b).
Attention computes $\mathrm{softmax}(QK^\top\!/\sqrt{d})\,V$ in three steps:
\begin{enumerate}
    \item \textbf{Pairwise comparison} ($QK^\top$): identifies which positions match.
    \item \textbf{Normalization} (softmax): concentrates weight on the best matches.
    \item \textbf{Value aggregation} ($\cdot V$): writes the match results back into each position's representation.
\end{enumerate}

Contrastive pretraining replicates step 1 --- it provides pairwise comparison supervision --- but not step 3 (value aggregation, which writes match results into representations).
This explains the failure: contrastive loss teaches which tokens match but does not update representations to encode these matches.
The match result must become part of the representation, not something the router must independently discover.

The deeper principle, validated by our escape routes (Section~\ref{sec:escape}): routing requires relational information to be \emph{written into} per-token representations.
The writing mechanism can be value aggregation (attention), sequential state propagation (bidirectional Mamba, 99.5\%), or inducing-point cross-attention (Perceiver hubs, 99.5\%) --- but contrastive pretraining, which only trains projections on \emph{unmodified} representations, cannot provide it.

Having established the mechanism on our synthetic task, we verify generalization before presenting the escape routes.

\subsection{Generalization}

\begin{figure}[t]
    \centering
    \includegraphics[width=0.9\textwidth]{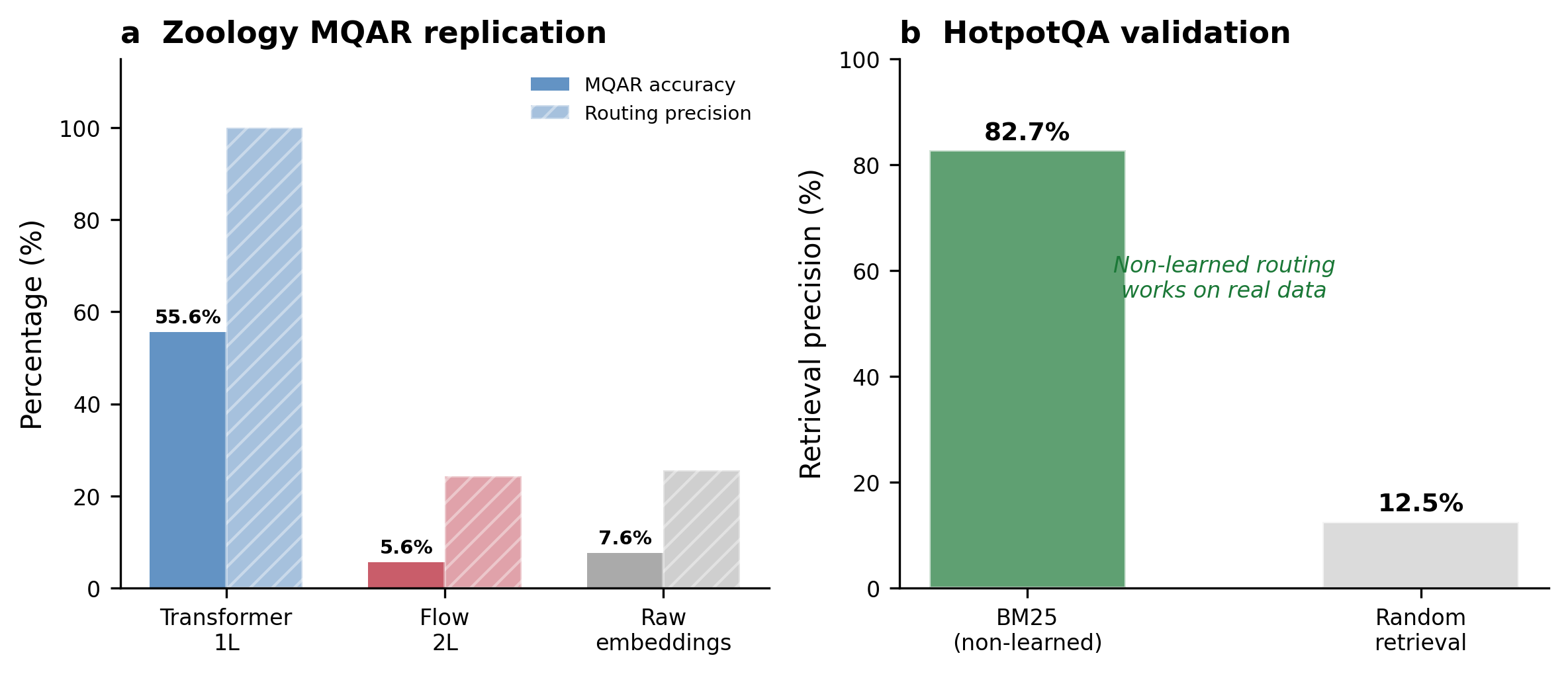}
    \caption{\textbf{Generalization beyond the synthetic task.} (a)~Our findings replicate on the Zoology MQAR benchmark: Transformer 1L achieves 100\% routing and 55.6\% accuracy; Flow and raw embeddings cluster at $\sim$25\%. (b)~BM25 achieves 82.7\% retrieval on HotpotQA with zero learned parameters.}
    \label{fig:generalization}
\end{figure}

Our findings generalize across tasks (Figure~\ref{fig:generalization}, Table~\ref{tab:generalization}).
On MQAR~\cite{arora2024zoology}: Transformer achieves 100\% routing (phase transition at epoch 5); Flow and raw embeddings cluster at $\sim$25\%.
On HotpotQA~\cite{yang2018hotpotqa}: BM25 keyword retrieval achieves 82.7\% sentence-segment precision with zero learned parameters, compared to a random-retrieval baseline of $\sim$12.5\% (selecting 2 of 16 segments at random).

\begin{table}[h]
\centering
\caption{\textbf{Generalization results.} The routing hierarchy replicates across tasks. MQAR routing precision = Investigator selects the correct key position. HotpotQA precision = at least one supporting sentence appears in retrieved segments.}
\label{tab:generalization}
\begin{tabular}{llcc}
\toprule
\textbf{Task} & \textbf{Model / Method} & \textbf{Routing / Retrieval} & \textbf{Task Acc} \\
\midrule
\multirow{3}{*}{MQAR~\cite{arora2024zoology}} & Transformer 1L & 100.0\% & 55.6\% \\
& Flow 2L & 24.2\% & 5.6\% \\
& Raw embeddings & 25.6\% & 7.6\% \\
\midrule
\multirow{2}{*}{HotpotQA~\cite{yang2018hotpotqa}} & BM25 (non-learned) & 82.7\% & ---\textsuperscript{$\dagger$} \\
& Random retrieval & 12.5\% & ---\textsuperscript{$\dagger$} \\
\bottomrule
\end{tabular}

\vspace{2pt}
{\small \textsuperscript{$\dagger$}HotpotQA evaluates retrieval precision only; end-to-end QA accuracy is not measured.}
\end{table}

These results establish our latent-subspace mechanism as the explanation for the recall gap observed by Arora et al.~\cite{arora2024zoology}.

Having confirmed the two-ingredient principle across tasks, we now ask: must the preprocessing ingredient be expensive attention, or can cheaper mechanisms provide the same bidirectional per-token representations?

\subsection{Escape Routes: $O(n)$ Alternatives to Attention}
\label{sec:escape}

If pairwise comparison is inescapable, can the \emph{preprocessing} be made cheaper?
We test six $O(n)$ or $O(n \log n)$ alternatives to bidirectional attention, all combined with the same pairwise Investigator router:

\vspace{4pt}
\begin{center}
\begin{tabular}{llcc}
\toprule
\textbf{Preprocessing} & \textbf{Cost} & \textbf{Routing} & \textbf{Acc} \\
\midrule
\multicolumn{4}{l}{\emph{Working ($>$98\%):}} \\
Bidir Mamba + rank-1 router & $O(n)$ & \textbf{99.7\%} & 30.6\% \\
Bidir Mamba & $O(n)$ & 99.5\% & 32.2\% \\
Perceiver $M{=}16$ hubs & $O(16n)$ & 99.5\% & 73.0\% \\
Perceiver $M{=}1$ hub & $O(n)$ & 98.8\% & 63.5\% \\
\midrule
\multicolumn{4}{l}{\emph{Failing ($<$2\%):}} \\
Global mean + broadcast & $O(n)$ & 1.9\% & 1.0\% \\
Fourier (FFT) mixing & $O(n \log n)$ & 0.9\% & 1.8\% \\
\midrule
\multicolumn{4}{l}{\emph{Controls:}} \\
Bidir attention 1L & $O(n^2)$ & 98.9\% & 75.6\% \\
Causal attention 1L & $O(n^2)$ & 99.2\% & 72.6\% \\
Unidir Flow & $O(n)$ & 2.2\% & 1.8\% \\
\bottomrule
\end{tabular}
\end{center}
\vspace{4pt}

Four of six alternatives achieve $>$98\% routing, matching or exceeding bidirectional attention.
Two fail completely.
The top entry --- ``Bidir Mamba + rank-1 router'' (99.7\%) --- replaces the full Investigator ($O(n^2)$ pairwise scoring across all position pairs) with a rank-1 pairwise projection: each token gets a scalar fingerprint $s_i = v^\top h_i$, and routing scores are $s_i \cdot s_j$ (rank-1 in the query-key product).
This reduces inference-time routing from $O(n^2)$ to $O(n)$ while achieving the highest routing precision of any method.
The ``Bidir Mamba'' entry (99.5\%) uses the full Investigator for routing.
The pattern is clear: preprocessing must produce \emph{per-token} representations with bidirectional context.

\paragraph{Why bidirectional Mamba works.}
Forward and backward scans give each position context from both directions.
Unlike unidirectional recurrence (where position $q$ has no information about future position $a > q$), bidirectional Mamba's backward pass propagates information from $a$ to $q$.
The Investigator's pairwise comparison can then detect the mutual context in both representations.

\paragraph{Why Perceiver hubs work.}
Learned inducing points cross-attend to \emph{all} tokens, forming a global summary, then broadcast back.
This gives each token access to global context through the hub relay --- analogous to the thalamic relay in neuroscience.
A single hub ($M{=}1$) suffices for 98.8\% routing, consistent with the rank-1 finding: one dimension of bidirectional information captures nearly all routing signal.

\paragraph{Why global mean and Fourier fail.}
Global mean pooling collapses all tokens into a single vector, destroying per-token discriminability.
Fourier mixing applies a global spectral transform but does not create \emph{relational} structure --- it preserves frequency information, not token-pair relevance.

\paragraph{Routing versus reasoning.}
An important dissociation: bidirectional Mamba achieves 99.5\% routing but only 32.2\% task accuracy, while Perceiver $M{=}16$ achieves the same routing at 73.0\% accuracy.
Both route correctly, but Perceiver's cross-attention produces richer representations that the Council can reason over.
Routing needs only bidirectional context (even compressed); reasoning benefits from richer relational structure.

\subsection{Scale and the Pairwise--Compression Divide}
\label{sec:scale}

Do our findings hold at scale? We test pre-trained models (370M--1.4B parameters), a bidirectional bridge experiment, and a controlled causal-vs-bidirectional ablation.

\begin{table}[h]
\centering
\caption{\textbf{The pairwise--compression divide.} Every system with pairwise token comparison achieves high routing. Every system without it fails. The divide is not causal-vs-bidirectional but pairwise-vs-compressed. The Investigator's dot-product scoring ($s_{ij} = q_i^\top k_j$) is itself pairwise computation --- systems with attention preprocessing AND pairwise routing have pairwise computation at two stages.}
\label{tab:scale}
\begin{tabular}{llccl}
\toprule
\textbf{Model} & \textbf{Type} & \textbf{Routing} & \textbf{Rank} & \textbf{Pairwise stage} \\
\midrule
\multicolumn{5}{l}{\emph{Systems with pairwise computation (all succeed):}} \\
Bidir attention 1L (E2E) & Bidir attn + router & 99.1\% & 1.6 & Preproc + router \\
Causal attention 1L (E2E) & Causal attn + router & \textbf{98.9\%} & 1.6 & Router only\textsuperscript{$\star$} \\
Sparse bidir (w32+g16) & Sparse bidir + router & 99.7\% & 1.5 & Preproc + router \\
Rank-1 bidir projection & Rank-1 bidir + router & 98.8\% & 1.6 & Preproc + router \\
Pythia-1B + bidir bridge & Frozen + bidir layer & \textbf{99.4\%} & \textbf{3.5} & Bridge + router \\
\midrule
\multicolumn{5}{l}{\emph{Systems without effective pairwise computation (all fail):}} \\
Pythia-1B (frozen) & Causal attn, no E2E & 35.0\% & 37.8 & Router only\textsuperscript{$\dagger$} \\
Mamba-1.4B (frozen) & Recurrence, no E2E & 28.7\% & 43.8 & Router only\textsuperscript{$\dagger$} \\
Mamba-370M (frozen) & Recurrence, no E2E & 22.5\% & 57.4 & Router only\textsuperscript{$\dagger$} \\
Flow 2L (E2E) & Unidir recurrence & $\leq 2\%$ & $>130$ & Router\textsuperscript{$\ddagger$} \\
Bidir Flow + seg.\ routing & Bidir rec., no pairwise & 25.6\% & --- & Segment\textsuperscript{$\S$} \\
No attention (E2E) & Raw embeddings & 1.2\% & 240 & Router\textsuperscript{$\|$} \\
\bottomrule
\end{tabular}

\vspace{2pt}
{\small \textsuperscript{$\star$}Causal attn preserves per-token repr.; router's pairwise comparison + E2E gradients suffice.\\
\textsuperscript{$\dagger$}Router has pairwise comparison but frozen repr.\ not shaped for routing.\\
\textsuperscript{$\ddagger$}Unidir recurrence: earlier positions lack future context, even with pairwise router.\\
\textsuperscript{$\S$}Bidir repr.\ are good (99.5\% with pairwise router), but segment-level routing lacks pairwise precision.\\
\textsuperscript{$\|$}No preprocessing: raw embeddings lack any contextual information.}
\end{table}

\paragraph{The key finding: pairwise computation, not directionality.}
In a controlled ablation --- same model, same training, only the attention mask differs --- causal attention matches bidirectional at \emph{every} depth:

\vspace{4pt}
\begin{center}
\begin{tabular}{lcccc}
\toprule
 & \textbf{1 layer} & \textbf{2 layers} & \textbf{4 layers} \\
\midrule
Bidirectional & 99.1\% & 99.1\% & --- \\
Causal & 98.9\% & 99.1\% & 99.1\% \\
\bottomrule
\end{tabular}
\end{center}
\vspace{4pt}

This seems to contradict a ``bidirectionality bottleneck'' narrative, but it actually \emph{confirms} the deeper principle.
The Investigator computes $s_{ij} = (W_q x_i)^\top (W_k x_j)$ across all position pairs --- this IS pairwise computation.
The causal model succeeds not because it escapes pairwise computation, but because the router \emph{provides} it.

The critical distinction is \textbf{bidirectional per-token representations versus unidirectional compression}:
\begin{itemize}
    \item \textbf{Bidirectional preprocessing} (attention or bidir recurrence) outputs one $d$-dimensional vector per token with context from both directions. The router can meaningfully compare $x_i$ to $x_j$. Bidirectional Mamba achieves 99.5\% routing; causal attention achieves 98.9\%.
    \item \textbf{Unidirectional recurrence} compresses positions $1 \ldots t$ into a fixed-size state. Position $q$'s representation has no information about future position $a > q$, making the comparison $s_{qa}$ indiscriminate. Result: $\leq 2\%$ routing even with pairwise comparison in the router.
\end{itemize}

\paragraph{The bidirectional bridge: unlocking frozen representations.}
Adding a single bidirectional attention layer on top of frozen Pythia-1B representations recovers \textbf{99.4\% routing} (rank 3.5), compared to 35\% without --- matching the small-model bidirectional baseline (99.1\%) despite the backbone being 2000$\times$ larger and frozen.
This confirms: pre-trained causal representations already contain the information routing needs, but it is encoded in a form inaccessible without pairwise transformation.
One bidirectional layer provides this transformation --- it allows position $q$ to ``see'' position $a$ and vice versa, creating the symmetric match signal that the Investigator's pairwise comparison requires.

\paragraph{Rank-1 pairwise computation suffices.}
Rank-1 bidirectional projection achieves 98.8\% routing.
The routing signal is essentially one-dimensional.
Although our implementation materializes the full $O(n^2)$ score matrix during training (for the auxiliary routing loss), the rank-1 structure means inference-time routing can be computed in $O(n)$: compute a scalar fingerprint $s_i = v^\top h_i$ per token, then select top-$k$ by query-key fingerprint product.

\paragraph{Implications for LLMs, Mamba, and hybrid architectures.}

\begin{itemize}
    \item \textbf{Mamba can preprocess but not route.}
    Unidirectional recurrence fails at routing regardless of training regime ($\leq 2\%$ end-to-end, 22--29\% pre-trained).
    But \emph{bidirectional} recurrence produces per-token representations that, combined with pairwise routing (the Investigator), achieve 99.5\%.
    The nuance: bidirectional Mamba serves as a cheap $O(n)$ \emph{preprocessor} that replaces bidirectional attention, but the pairwise routing step ($O(n^2)$) remains necessary.
    Without pairwise routing (segment-level routing only), bidirectional Mamba drops to 25.6\%.
    The practical implication: in hybrid architectures, SSM layers can handle representation building, but the routing decision itself requires pairwise token comparison.

    \item \textbf{Why MoE gating works.} MoE gates~\cite{fedus2022switch,jiang2024mixtral} are trained end-to-end on representations from \emph{attention} layers that preserve per-token identity. Our causal-attention result (98.9\%) explains their success: attention-based representations + end-to-end routing supervision suffice.

    \item \textbf{Why BERT retrieval beats LLM retrieval.} Dense retrieval on frozen causal LLM features underperforms BERT-based bi-encoders~\cite{karpukhin2020dense}. Our results explain why: frozen causal representations lack routing-compatible structure ($\sim$35\%). BERT provides bidirectional pairwise preprocessing that creates it. BM25~\cite{robertson2009probabilistic} remains competitive because it bypasses learned representations entirely.

    \item \textbf{Hybrid architectures: SSM layers preprocess, attention layers route.} Jamba~\cite{lieber2024jamba}, Griffin~\cite{de2024griffin}, and Zamba~\cite{glorioso2024zamba} interleave attention with recurrence. Our results predict a division of labor: SSM layers can cheaply build contextualized per-token representations (as bidirectional Mamba demonstrates), while attention layers provide the pairwise computation routing requires. Routing on SSM-layer representations alone fails without pairwise comparison.

    \item \textbf{The practical prescription.} Pairwise token comparison is inescapable for routing. The preprocessing can be cheap ($O(n)$ bidirectional recurrence suffices). The pairwise comparison itself can also be cheap (rank-1 at $O(n)$ inference, sparse at $O(n \cdot w)$). But it cannot be zero. Pure recurrence without any pairwise stage is a dead end.
\end{itemize}

%----------------------------------------------------------------------
\section{Discussion}
\label{sec:discussion}
%----------------------------------------------------------------------

\paragraph{The routing quality hierarchy.}
Our experiments reveal a sharp divide (Table~\ref{tab:hierarchy}).

\begin{table}[h]
\centering
\caption{\textbf{The routing quality hierarchy.} Routing requires two ingredients: (1)~per-token representations with bidirectional context, and (2)~pairwise token comparison. The preprocessing can be attention, bidirectional recurrence, or inducing points --- but pairwise comparison cannot be absent. Combining $O(n)$ bidir Mamba with rank-1 pairwise yields the best routing (99.7\%) at linear cost.}
\label{tab:hierarchy}
\begin{tabular}{llcl}
\toprule
\textbf{Regime} & \textbf{Routing} & \textbf{Ingredient(s) present} & \textbf{Cost} \\
\midrule
\multicolumn{4}{l}{\emph{Both ingredients present (all succeed):}} \\
Bidir Mamba + rank-1 pairwise & \textbf{99.7\%} & Bidir preproc + pairwise & $O(n)$ \\
Sparse bidir attn (w32+g16) & 99.7\% & Bidir preproc + pairwise & $O(n \cdot w)$ \\
Perceiver $M{=}16$ + pairwise & 99.5\% & Inducing pts + pairwise & $O(16n)$ \\
Pythia-1B + bidir bridge & 99.4\% & Bridge + pairwise & $O(n^2)$ \\
Bidir attn 1L + pairwise & 99.1\% & Bidir attn + pairwise & $O(n^2)$ \\
Causal attn 1L (E2E) + pairwise & 98.9\% & Causal attn + pairwise & $O(n^2)$ \\
Perceiver $M{=}1$ + pairwise & 98.8\% & 1 ind.\ pt + pairwise & $O(n)$ \\
Non-learned index & 82--91\% & Hash/BM25 (no learning) & $O(n)$ \\
\midrule
\multicolumn{4}{l}{\emph{One or both ingredients missing (all fail):}} \\
Pythia-1B frozen (no E2E) & 35\% & Per-token but not shaped & $O(n^2)$ \\
Mamba-1.4B frozen & 28.7\% & Compressed + post-hoc & $O(n)$ \\
Bidir Mamba + segment routing & 25.6\% & Bidir preproc, no pairwise & $O(n)$ \\
Unidir Flow + pairwise & 2.2\% & Pairwise, no bidir context & $O(n^2)$ \\
Global mean + pairwise & 1.9\% & Pairwise, no per-token repr & $O(n^2)$ \\
No preproc + pairwise & 1.2\% & Pairwise, no context & $O(n^2)$ \\
Fourier + pairwise & 0.9\% & Pairwise, no relational info & $O(n^2)$ \\
\bottomrule
\end{tabular}
\end{table}

Table~\ref{tab:hierarchy} consolidates all results into the \textbf{two-ingredient principle}: routing requires both (1)~per-token representations with bidirectional context, and (2)~pairwise token comparison.
Every entry with both ingredients achieves $>$98\%; every entry missing either one falls below 35\%.
The gap is not gradual --- it is a discrete divide.

\paragraph{The routing paradox and its resolution.}
Content-based routing requires pairwise token comparison.
This is not negotiable: 20+ experiments confirm that every non-pairwise mechanism fails.
But the cost can be driven far below $O(n^2)$: rank-1 projection ($O(n)$) and sparse patterns ($O(n \cdot w)$) both achieve $>$98\%.
Moreover, the \emph{preprocessing} can be cheap: bidirectional Mamba ($O(n)$) produces representations as good as bidirectional attention ($O(n^2)$) for the router --- achieving 99.5\% routing when combined with pairwise comparison.
The practical resolution: cheap $O(n)$ preprocessing (bidirectional recurrence) + cheap pairwise comparison (rank-1 or sparse) = effective routing at sub-quadratic cost (linear at inference time; training requires $O(n^2)$ for the auxiliary routing loss).

\paragraph{Unidirectional recurrence remains a dead end.}
Unidirectional recurrence cannot support routing even with pairwise comparison in the router (2.2\%).
The issue is asymmetric context: at position $t$, the recurrent state encodes positions $1 \ldots t$ but nothing about $t{+}1 \ldots n$.
The router compares position $q$ to position $a > q$, but $q$'s representation has no information about $a$, producing indiscriminate scores.
Bidirectional recurrence fixes this: position $q$'s backward pass has seen $a$, and $a$'s forward pass has seen $q$, so both representations carry mutual context.
This explains the dramatic gap: bidir Mamba + Investigator = 99.5\%, unidir Flow + Investigator = 2.2\%.

\paragraph{Connection to the recall--throughput tradeoff.}
Arora et al.~\cite{arora2024based} identified a recall--throughput tradeoff in efficient models: sub-quadratic models trade recall for speed.
Our work reveals the mechanism: recall requires per-token representations that pairwise comparison can discriminate.
Unidirectional recurrence trades recall for speed because earlier positions lack future context.
Bidirectional recurrence resolves this at $O(n)$ cost, but the pairwise comparison step remains necessary.
The tradeoff is not between model types but between the \textbf{presence or absence of both ingredients}: bidirectional per-token representations and pairwise comparison.

\paragraph{Scope: why synthetic-task findings transfer.}
Our core experiments use a controlled synthetic task (Distant Evidence Retrieval) to isolate routing requirements.
A natural concern: do findings on exact token matching at 486K parameters transfer to real language at scale?
Three lines of evidence support transferability.
First, the two-ingredient principle replicates on MQAR~\cite{arora2024zoology} (a standard benchmark) and HotpotQA (real-world QA with BM25).
Second, the Pythia-1B bridge experiment (99.4\% routing on a frozen 1B-parameter pre-trained model) validates the principle at realistic scale --- the routing failure on Pythia's frozen representations and its recovery with one bidirectional layer are not synthetic-task artifacts.
Third, the principle's predictions align with real-world patterns: MoE gating works (end-to-end pairwise on attention representations), BERT retrieval outperforms causal LM retrieval (bidirectional preprocessing), and BM25 remains competitive (non-learned bypass).
That said, we acknowledge that the synthetic task isolates exact token matching --- the simplest routing criterion.
Three open directions remain:
\emph{Semantic matching}: all methods struggle when matching requires meaning-level similarity rather than token identity. Our initial experiments (structured embedding initialization, anchor-based sponsorship) show 0.6--11\% routing, suggesting a possible third ingredient --- semantic structure in the representations --- that neither pairwise computation nor bidirectional context provides.
\emph{Cheaper pairwise comparison}: Perceiver-style inducing points already work ($M{=}16$: 99.5\%, $M{=}1$: 98.8\%), but whether learned hashing can eliminate pairwise computation entirely remains open.
\emph{Architecture design}: The two-ingredient principle prescribes a concrete architecture --- $O(n)$ bidirectional Mamba preprocessing + $O(n)$ rank-1 pairwise routing (99.7\% combined) + sparse Council reasoning --- that achieves linear-cost routing. Validating this design at scale on real benchmarks (SCROLLS, QuALITY) is natural follow-up work.

\paragraph{Limitations.}
Sequential Python scan limits SSM speed; HotpotQA validation uses only BM25 retrieval precision; contrastive pretraining tested only with InfoNCE.
We do not compare against modern sub-quadratic hybrids (Based~\cite{arora2024based}, GLA, HGRN2) which mix linear and window attention --- these methods include window attention (a form of local pairwise computation), which our principle predicts should enable routing within each window, consistent with our sparse attention result (99.7\%).
The bidirectional Mamba result (99.5\%) and causal attention result (98.9\%) both rely on the Investigator's full $O(n^2)$ pairwise comparison --- architectures with cheaper routing (e.g., MoE gating) may behave differently.
The Pythia bidirectional bridge (99.4\%) uses a full-attention bridge layer; sparse or rank-1 bridges remain untested at scale.
Adaptive budget allocation (MF-OFUL 50\%: 91.2\%) shows promise but exhibits training instability near the phase transition boundary, and is tested only on the synthetic task.
The ``$O(n)$ linear cost'' claim for rank-1 routing applies to inference; during training, the auxiliary routing loss requires the full $O(n^2)$ score matrix.

%----------------------------------------------------------------------
\section{Conclusion}
\label{sec:conclusion}
%----------------------------------------------------------------------

We exhaustively map the routing landscape for hybrid sequence models across 20+ controlled experiments spanning three tasks, nine representation types, four routing mechanisms, two training signals, and multiple scales (200K to 1.4B parameters).

The landscape is governed by a \textbf{two-ingredient principle}: content-based routing requires (1)~per-token representations with bidirectional context and (2)~pairwise token comparison in the router.
Either ingredient alone fails; both together succeed regardless of architecture.

The principle produces sharp, constructive predictions:
\begin{itemize}
    \item \textbf{Preprocessing can be cheap.} Six different $O(n)$ or $O(n \cdot m)$ mechanisms match $O(n^2)$ attention for routing: bidirectional Mamba (99.5\%), Perceiver $M{=}16$ (99.5\%), Perceiver $M{=}1$ (98.8\%), causal attention with E2E training (98.9\%), sparse attention (99.7\%), and rank-1 bidir projection (98.8\%). The preprocessing does not need to be attention.
    \item \textbf{Pairwise comparison is inescapable.} Every mechanism that avoids pairwise token comparison fails: segment-level routing drops from 99.5\% to 25.6\%; global mean pooling = 1.9\%; Fourier mixing = 0.9\%; memory banks = 12\%; bandits = 0.7--3.6\%. But pairwise comparison can be rank-1 ($O(n)$: 98.8\%) or sparse ($O(n \cdot w)$: 99.7\%).
    \item \textbf{The optimal combination is linear-cost.} Bidirectional Mamba ($O(n)$) + rank-1 pairwise ($O(n)$) achieves \textbf{99.7\% routing} --- the highest of any method --- at $O(n)$ total cost. This provides a concrete, actionable architecture prescription.
    \item \textbf{Unidirectional recurrence is a dead end.} Even with full pairwise comparison, unidirectional recurrence achieves only 2.2\%. Bidirectionality in preprocessing is necessary.
\end{itemize}

This reframes the recall gap~\cite{arora2024zoology,fu2023h3}: Transformers recall because they provide both ingredients (per-token representations + pairwise computation); SSMs provide neither (compressed state + no pairwise comparison).
Hybrid architectures~\cite{lieber2024jamba,de2024griffin,glorioso2024zamba} succeed by combining SSM layers (cheap preprocessing) with attention layers (pairwise computation).
The practical prescription: $O(n)$ bidirectional preprocessing + $O(n)$ rank-1 pairwise comparison = effective routing at \textbf{linear inference cost} (training still requires $O(n^2)$ for the auxiliary routing loss, but the trained model routes in $O(n)$).

\end{document}